\documentclass[10pt,twocolumn,letterpaper]{article}
\usepackage[pagenumbers]{cvpr}

\usepackage{multirow}
\usepackage{diagbox}
\usepackage{makecell} 
\usepackage{colortbl}
\usepackage{xcolor}
\usepackage{bbm}
\usepackage{subcaption}

\usepackage{soul}
\usepackage{wrapfig}
\usepackage{amsthm}
\usepackage{stmaryrd}

\definecolor{lightgray}{gray}{0.9}
\usepackage{pifont}
\newcommand{\cmark}{\ding{51}}%
\newcommand{\xmark}{\ding{55}}%

\usepackage{amsthm}

\newtheorem*{assumption*}{\textit{Assumption (anomaly scarcity)}}
\newtheorem*{hypothesis*}{\textit{Hypothesis (dominant cluster existence)}}

\definecolor{cvprblue}{rgb}{0.21,0.49,0.74}
\usepackage[pagebackref,breaklinks,colorlinks,allcolors=cvprblue]{hyperref}
\usepackage[capitalize]{cleveref}

\crefname{section}{Section}{Sections}
\Crefname{section}{Section}{Sections}
\crefname{table}{Table}{Tables}
\Crefname{table}{Table}{Tables}
\crefname{figure}{Figure}{Figures}
\Crefname{figure}{Figure}{Figures}
\crefname{equation}{Equation}{Equations}
\Crefname{equation}{Equation}{Equations}

\title{When Unsupervised Domain Adaptation meets One-class Anomaly Detection: Addressing the Two-fold Unsupervised Curse by Leveraging Anomaly Scarcity}

\author{
Nesryne Mejri$^{1}$, Enjie Ghorbel$^{2,1}$, Anis Kacem$^{1}$, Pavel Chernakov$^{1}$,\\ 
Niki Foteinopoulou$^{1}$, Djamila Aouada$^{1}$\\
$^{1}$Interdisciplinary Centre for Security, Reliability and Trust (SnT), University of Luxembourg\\
$^{2}$Cristal Laboratory, National School of Computer Sciences, University of Manouba\\
{\tt\small \{nesryne.mejri, anis.kacem, pavel.chernakov, niki.foteinopoulou, djamila.aouada\}@uni.lu}\\
{\tt\small 
enjie.ghorbel@isamm.uma.tn}
}

\begin{document}
\maketitle
\begin{abstract}
This paper introduces the first fully unsupervised domain adaptation (UDA) framework for unsupervised anomaly detection (UAD). The performance of UAD techniques degrades significantly in the presence of a domain shift, difficult to avoid in a real-world setting. While UDA has contributed to solving this issue in binary and multi-class classification, such a strategy is ill-posed in UAD. This might be explained by the unsupervised nature of the two tasks, namely, domain adaptation and anomaly detection. Herein, we first formulate this problem that we call \textbf{the two-fold unsupervised curse}. Then, we propose a pioneering solution to this curse, considered intractable so far, by assuming that anomalies are rare. Specifically, we leverage clustering techniques to identify a dominant cluster in the target feature space. Posed as the normal cluster, the latter is aligned with the source normal features. Concretely, given a one-class source set and an unlabeled target set composed mostly of normal data and some anomalies, we fit the source features within a hypersphere while jointly aligning them with the features of the dominant cluster from the target set. The paper provides extensive experiments and analysis on common adaptation benchmarks for anomaly detection, demonstrating the relevance of both the newly introduced paradigm and the proposed approach. \texttt{The code will be made publicly available.}

\end{abstract}    

\begin{figure}
    \centering
    \includegraphics[scale=0.32]{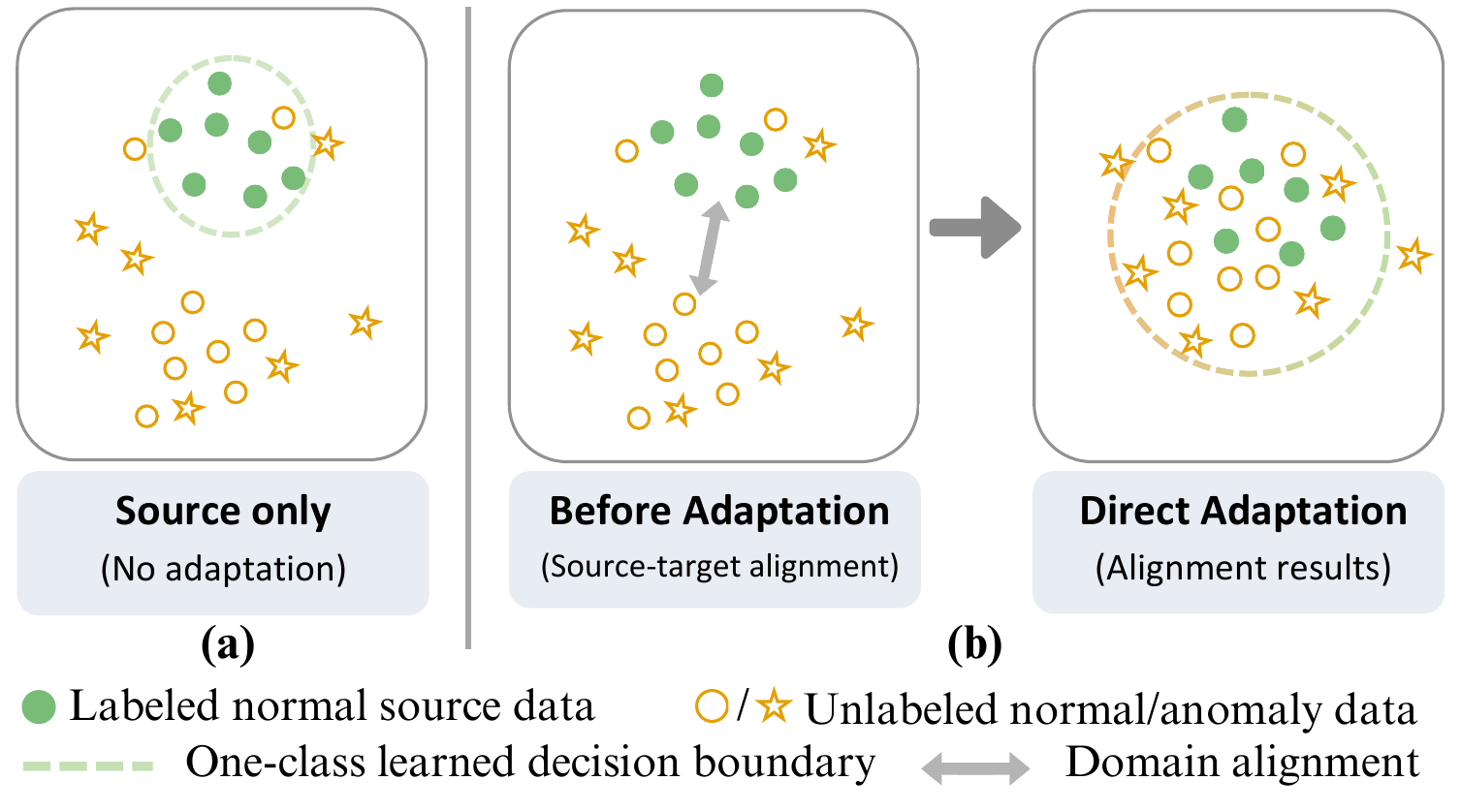}
    \vspace{-3mm}
    \caption{\textbf{Illustration of the two-fold unsupervised curse}: \textbf{(a)} The decision boundary learned from the source set without any adaptation does not allow generalization to the target domain. \textbf{(b)} Direct alignment of the unlabeled target with the one-class source features leads to the confusion of normal and abnormal samples.}
    \vspace{-3mm}
    \label{fig:double-unsupervision}
\end{figure}
\section{Introduction}
\label{sec:intro}
Anomaly Detection (AD) can be seen as the identification of outliers deviating from a usual pattern. The growing interest for AD in both academia and industry is mainly due to its relevance in numerous practical scenarios, such as early disease detection in medical imaging~\cite{medical-application-2024-2,medical-application2024} and industrial inspection~\cite{ctt-da-fink-2024,uad-ts-2024,rd4ad-2022,warehouse-application-2018, indus-survey-2024}. By definition, anomalies rarely occur. Annotating anomalous data is, therefore, often difficult and costly~\cite{romain-2025,ctt-da-fink-2024,ad-shift-2023}, hindering the collection of large-scale datasets. As a result, state-of-the-art methods mostly tackle AD as an unsupervised problem~\cite{ad-bench-2022,ruff-survey-2021}, where the objective is to learn from the normal class.

Despite achieving promising results, recent approaches in AD~\cite{dsvdd-2018,rd4ad-2022,rd4ad++-2023,semantic-ad-2024,romain-2025} typically assume that training and inference data are drawn from the same distribution. This assumption does not always hold in unconstrained scenarios, where a \textit{domain shift}~\cite{domain-shift} between training and testing data can naturally arise due to varying setups, such as different lighting conditions and variations in object pose~\cite{ad-shift-2023}.  As a result, a model trained on a dataset sampled from a given domain, usually called \textit{source} dataset, will show degraded performance when tested on a dataset from a different domain, generally termed \textit{target} dataset. For instance, an AD model for medical imaging trained on images acquired using a given Medical Resonance Imaging (MRI) device can fail to generalize to samples captured with a different MRI system. 

\begin{figure}[t!]
    \centering
    \includegraphics[scale=0.325]{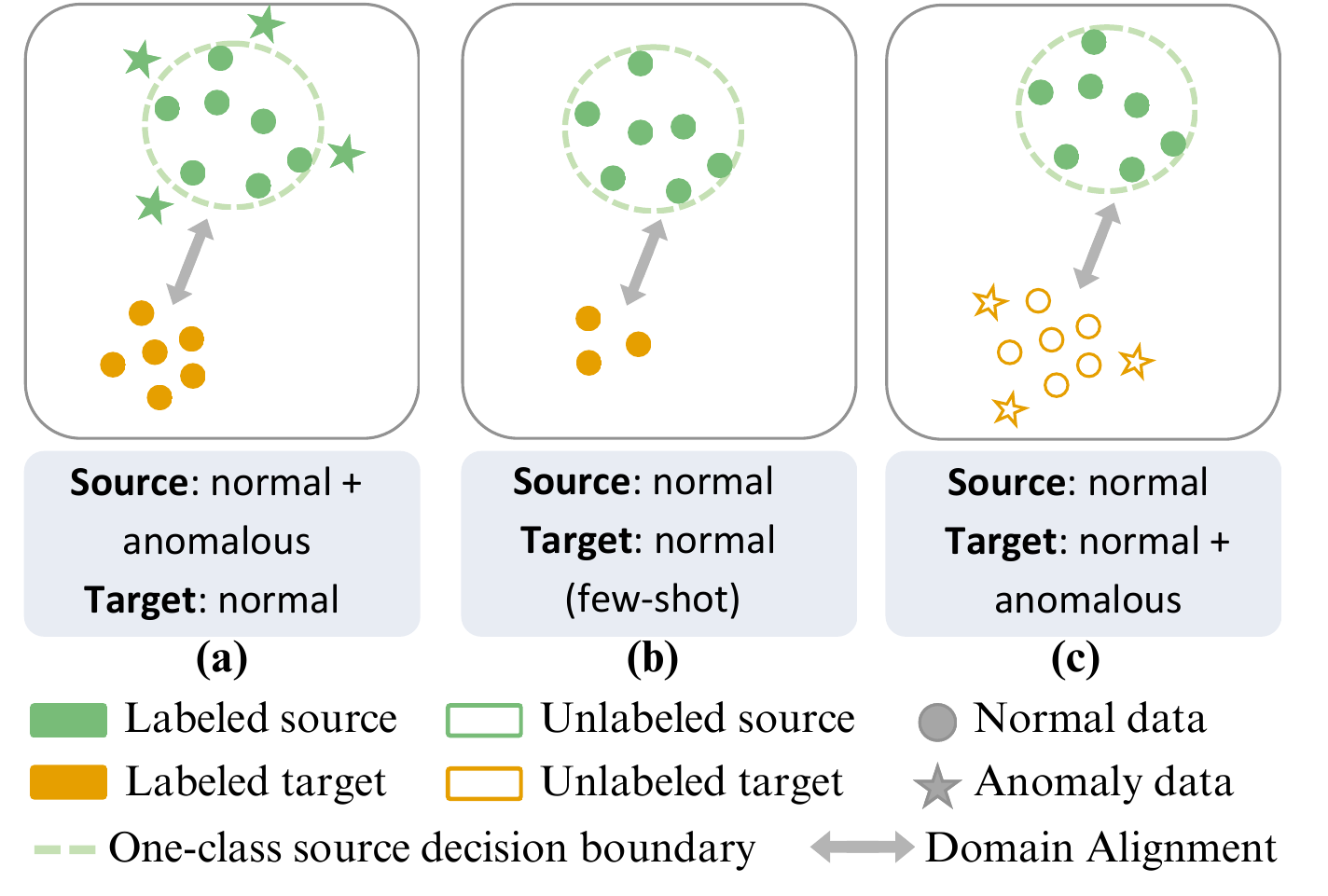}
    \vspace{-2mm}
    \caption{\textbf{Comparison of our setting with previous works}: \textbf{(a)} supervised source anomaly detection with supervised domain adaptation~\cite{ildr-2019}, \textbf{(b)} unsupervised one-class source anomaly detection with few-shot domain adaptation~\cite{msra-2021,irad-2023,iris-attack-detection-2022}, \textbf{(c)} \textbf{our considered setting}: unsupervised one-class source anomaly detection with unsupervised domain adaptation.}
    \label{fig:teaser}
    \vspace{-5mm}
\end{figure}

To reduce such a domain gap while avoiding costly annotation efforts, Unsupervised Domain Adaptation (UDA)~\cite{uda-survey-2020,uda-bench-2024} has proven to be an effective solution in binary and multi-class classification tasks~\cite{inder-wacv-2024,uda-bench-2024}. UDA aims at learning domain-invariant features by relying on labeled source and unlabeled target data at the same time. However, the task of \textbf{unsupervised} domain adaptation for \textbf{unsupervised} anomaly detection (UAD) is ill-posed as the goal is to: \textit{align the source and the target feature distributions using only normal source data and unlabeled target data formed by both normal and anomalous samples} (see \Cref{fig:teaser} (c)). Hence, a direct extension of standard UDA techniques developed for binary/multi-class classification~\cite{uda-bench-2024, uda-survey-2020} would not be applicable, as these methods usually aim at minimizing the distance between the estimated distributions from the entire source and target training sets. Indeed, this would lead to the erroneous alignment of both normal and anomalous target samples with normal source samples, as illustrated in~\Cref{fig:double-unsupervision}~(b). Given the learned decision boundary, this would lead to the confusion of normal and abnormal samples from the target set. As it requires addressing two unsupervised tasks simultaneously, we refer to this described problem as the \textit{two-fold unsupervised curse}. 

To the best of our knowledge, no prior work has tried to address this two-fold unsupervised challenge, \ie, UDA for one-class image anomaly detection described in~\Cref{fig:teaser} (c). Indeed, related works have mainly simplified the problem by either (1) assuming the availability of labeled abnormal and normal source data, resulting in UDA for a binary classification setting~\cite{ildr-2019} (see~\Cref{fig:teaser} (a)), or (2) maintaining the source one-class setup while accessing only few normal target data referred to as few-shot supervised adaptation for unsupervised anomaly detection~\cite{ildr-2019,biost-2019,msra-2021,iris-attack-detection-2022,irad-2023} (see~\Cref{fig:teaser} (b)). Nevertheless, annotating a few samples might still be constraining, particularly in the field of anomaly detection, where expert knowledge is often needed, such as for tumor annotation in medical images~\cite{medical-application-2024-2,medical-application2024} or for industrial inspection~\cite{rd4ad-2022,rd4ad++-2023,indus-survey-2024}. Moreover, few-shot approaches are known to be prone to overfitting issues since few shots cannot fully represent the normal target distribution~\cite{challenges-few-shot-learning-2023}. This calls for a fully unsupervised domain adaptation approach that leverages the diversity of the available large unlabeled target datasets.

In this paper, we propose solving the two-fold unsupervised curse by leveraging the fact that the occurrence of anomalies tends to be rare. We herein propose the first unsupervised domain adaptation framework for unsupervised image anomaly detection. 
Our solution starts by identifying a dominant cluster assumed to be formed by normal target data and then aligning it with normal source samples. Specifically, our method utilizes a trainable ResNet-based~\cite{resnet-2016} feature extractor to process both the source and target features. A frozen CLIP visual encoder~\cite{clip-2021} is also used to generate corresponding target features, which are then clustered using K-means to identify the samples of the dominant cluster. These samples are mapped into the ResNet-based~\cite{resnet-2016} feature space and aligned with the source features.  For the domain adaptation task, a contrastive strategy~\cite{clip-2021,infonce-loss-2018} ensures the similarity between the dominant target cluster and normal source samples, while for the anomaly detection task, a Deep Support Vector Data Description (DSVDD)~\cite{dsvdd-2018} objective enforces feature compactness on the normal source data. 
Our framework is modular, allowing for flexible component changes, and supports various adaptation strategies, including statistical and adversarial alignment.  Experiments performed on standard UDA  benchmarks~\cite{office31,officehome,visda-2017} for semantic anomaly detection~\cite{semantic-ad-2024} demonstrate its effectiveness. Our approach achieves state-of-the-art (SoA) performance, even compared to few-shot methods.

\noindent\textbf{Contributions.} The main contributions of this work can be summarized as follows:
\begin{itemize}
    \item The two-fold unsupervised curse of UDA for one-class anomaly detection is formalized, and the induced challenges are outlined. 
    \item A solution to the two-fold unsupervised problem is proposed by leveraging an intrinsic property of anomalies, \ie, their scarcity.
    \item A UDA method for one-class anomaly detection is introduced, leveraging a Vision Language Model, namely CLIP~\cite{clip-2021}, for dominant cluster identification and alignment using a contrastive strategy.
    \item Extensive experiments and analysis are conducted on several benchmarks~\cite{office31,officehome,visda-2017,pacs-2017}, demonstrating the relevance of the proposed framework under both fully unsupervised and few-shot settings.
\end{itemize}
\textbf{Paper Organization.} \Cref{sec:related_works} reviews UAD works under domain shift. \Cref{sec:problem_formulation} defines the two-fold unsupervised curse, while \Cref{sec:generic_framework} and \Cref{sec:proposed_method} detail the proposed solution for solving it. \Cref{sec:experiments} and~\Cref{sec:limitations} cover the experiments and limitations of this method. \Cref{sec:conclusion} concludes and outlines future work.

\section{Related Works: Anomaly detection under domain shift}
\label{sec:related_works}

Unsupervised image anomaly detection is a well-established research area~\cite{ad-bench-2022,ruff-survey-2021,romain-2025,dsvdd-2018,rd4ad-2022,rd4ad++-2023,semantic-ad-2024} where the aim is to learn a function $\zeta$ using a single class corresponding to normal data from the normal-only dataset $\mathcal{D}^{n} =\{(\mathbf{X}_i, y_i); y_i=0\}_{i=1}^{N}$, to classify whether an input image $\mathbf{X}$ is normal ($y=0$) or not ($y=1$). This is achieved by optimizing the following objective,
\vspace{-1mm}
\begin{equation}
\min_{\zeta} \, \mathbb{E}_{(\mathbf{X}_i,y_i) \sim \mathcal{D}^n} \left[ \mathcal{L}\left(\zeta(\mathbf{X}_i), y_i=0\right) \right],
\vspace{-1mm}
\end{equation}
where $\mathcal{L}$ is a loss enforcing feature compactness as in DSVDD~\cite{dsvdd-2018} or a reconstruction loss typically used in autoencoders-based methods~\cite{rd4ad-2022,rd4ad++-2023}.
Although achieving impressive performance on standard benchmarks, the majority of AD methods~\cite{ad-bench-2022,ruff-survey-2021,romain-2025,dsvdd-2018,rd4ad-2022,rd4ad++-2023,semantic-ad-2024} overlook the domain gap problem where training and testing data denoted as $\mathcal{D}^{s}$ and $\mathcal{D}^{t}$, respectively, follow different distributions due to uncontrolled variations in the acquisition setting~\cite{ad-shift-2023,ad-shift-neurips-2024}. This domain shift induces, therefore, a significant drop in performance. To solve this issue, a handful of Domain Generalization~(DG) methods for UAD has been proposed recently~\cite{red-pandas,ad-shift-neurips-2024,ad-shift-2023}. \textcolor{black}{Cohen et al.~\cite{red-pandas} propose a domain-disentanglement approach that removes predefined nuisance attributes (e.g., pose, lighting) from the source features using contrastive loss, preventing these factors from interfering with the anomaly task, improving the performance on unseen domains. However, without an actual target set, this method requires defining and labeling nuisance factors within the source dataset, which is challenging, as mentioned in their paper.} In~\cite{ad-shift-neurips-2024}, multiple source domains are considered for learning domain-invariant features, thereby assuming the availability of diverse large-scale datasets, which is not always guaranteed. To avoid relying on multiple domains during training, a self-supervised strategy is adopted in~\cite{ad-shift-2023}. Nevertheless, the success of this approach heavily depends on the similarity between the augmented data and target samples. As a result, it necessitates tailoring augmentation techniques to unseen target datasets, if at all possible. 
Given its effectiveness, domain adaptation has also been explored to address the domain shift problem in AD~\cite{biost-2019, tsa-2021, ildr-2019,msra-2021,iris-attack-detection-2022}. Those techniques usually adopt a few-shot adaptation paradigm by having access to a limited number of annotated target samples. While these methods offer innovative solutions for aligning source and target normal data, they still rely on costly annotations~\cite{romain-2025} and are exposed to overfitting risks~\cite{challenges-few-shot-learning-2023}.  This emphasizes the need for a fully unsupervised domain adaptation for UAD. However, solving this challenge remains challenging given the doubly unsupervised nature of the problem resulting from both UAD and UDA, which is further described in the next section.

\section{The Two-fold Unsupervised Curse}

\label{sec:problem_formulation}

Let us denote as $\mathcal{D}^s = \{(\mathbf X_i^s, y_i^s)\}_{i=1}^{N_s}$ a labeled dataset from a given domain called \textit{source} formed by $N_s$ samples, where a sample $\mathbf X_i^s \in \mathbb{R}^{h \times w \times c}$ and its associated label $y_i^s \in \{0, 1\}$,  $\forall i=\{1,...,N_s\}$. Let $\mathcal{D}^t$ be a second unlabeled dataset from a different domain, \ie, \textit{target}, denoted as $\mathcal{D}^t = \{\mathbf X_i^t\}_{i=1}^{N_t}$ and formed by $N_t$ samples where $\mathbf X_i^t \in \mathbb{R}^{h \times w \times c}$,  $\forall  i=\{1,...,N_t\}$. In the following, we assume that $\mathcal{D}^t$  shares the same label space as $\mathcal{D}^s$ and that there exists a domain gap between $\mathcal{D}^s$ and $\mathcal{D}^t$. The goal of Unsupervised Domain Adaptation (UDA) for anomaly detection (whether formulated as a binary or one-class classification problem),  is to learn a model $ \zeta : \mathbb{R}^{h \times w \times c} \rightarrow \{0, 1\}$ using both $\mathcal{D}^s$ and $\mathcal{D}^t$ that generalizes to the target domain. In other words, it aims at learning a domain invariant feature extractor $f: \mathbb{R}^{h \times w \times c} \mapsto \mathcal{X} $ such that $\zeta =g \circ f $ with $g: \mathcal{X} \mapsto \{0, 1\} $ being the classifier and $\mathcal{X}$ the feature space given by $f$. This objective is achieved by minimizing the following adaptation upper bound~\cite{da-gen-bound-2006},
\begin{equation}
\label{eq:adaptation_bound}
\epsilon^t \leq \epsilon^s + d(f(\mathcal{D}^s), f(\mathcal{D}^t)) + \lambda \ ,
\end{equation}
where $\epsilon_t$ and $\epsilon_s$ are the expected classification errors on the target and source domains, respectively; $d(f(\mathcal{D}^s), f(\mathcal{D}^t))$ estimates the discrepancy between the feature distributions from the two domains, and $\lambda$ accounts for the error of an ideal detector. 

While strategies for minimizing this upper bound are feasible in the context of binary or even multi-class classification~\cite{uda-bench-2024,inder-wacv-2024,uda-survey-2020}, the non-availability of anomalous data during training makes it difficult in the context of one-class classification, where  $d(f(\mathcal{D}^s), f(\mathcal{D}^t))$ cannot be estimated. In fact, we can only use a subset $\mathcal{D}^{s,n} \subset \mathcal{D}^{s}$ formed by normal data for training. For that reason, existing works on domain adaptation for one-class anomaly detection~\cite{biost-2019,msra-2021,irad-2023} revisit the formulation given in Eq~\eqref{eq:adaptation_bound} by slightly simplifying the problem. They pose it as a few-shot domain adaptation setting (instead of a fully unsupervised scenario). This means that they assume having access to a small subset $\mathcal{D}^{t,n} \subset \mathcal{D}^{t}$ composed of normal samples only. As a result, they reformulate Eq~\eqref{eq:adaptation_bound} as follows,
\begin{equation}
    \label{eq:previous_works_uda_bound}
    \epsilon^{t,n}\leq \epsilon^{s,n} + d(f(\mathcal{D}^{s,n}), f(\mathcal{D}^{t,n})) + \lambda \ .
\end{equation}

\noindent where $\epsilon^{s,n}$ and $\epsilon^{t,n}$ represent the source and target expected classification errors related to the normal class, respectively, since $\epsilon^{s}$ is not measurable in this context.

Nevertheless, in a fully unsupervised setting, we have access to $\mathcal{D}^{t}=\mathcal{D}^{t,a} \cup \mathcal{D}^{t,n} $ where $\mathcal{D}^{t,a}$ represents the subset of $\mathcal{D}^{t}$ formed by anomalies, without any prior information regarding the labels. Hence, directly aligning the feature distributions estimated from the source and target data by approximating $d(f(\mathcal{D}^{s,n}), f(\mathcal{D}^{t}))$ would lead to obtaining a classification boundary that is completely obsolete for target data, as shown in Figure~\ref{fig:double-unsupervision} (b). We call this problem the \textit{two-fold unsupervised curse} as it is a consequence of a lack of supervision: (1) in the task of anomaly detection, as it is formulated as a one-class problem where only normal source data are used; and (2) in the task of domain adaptation which is fully unsupervised where only an unlabeled target set is available. Given that the problem is ill-posed, it remains a significant challenge that has not been addressed in the existing UAD literature.

\section{Rare Anomalies to the Rescue}
\label{sec:generic_framework}
To tackle the two-fold unsupervised curse described in Section~\ref{sec:problem_formulation}, we introduce a key assumption and the main hypothesis it entails for enabling unsupervised domain adaptation for one-class anomaly detection.

\vspace{-1mm}
\begin{assumption*}
For an unlabeled target dataset $\mathcal{D}^t=\mathcal{D}^{t,n} \cup \mathcal{D}^{t,a}$, we assume that the number of anomalous samples is significantly smaller than the number of normal samples, i.e., $|\mathcal{D}^{a}| << |\mathcal{D}^{n}|$, with $|.|$ refers to the cardinality.\\
\end{assumption*}
\vspace{-6mm}
\begin{hypothesis*}
Considering a target unlabeled anomaly detection dataset $\mathcal{D}^t=\mathcal{D}^{t,n} \cup \mathcal{D}^{t,a}$ \textbf{under the anomaly scarcity assumption}, where $\mathcal{D}^{t,n}$ and $\mathcal{D}^{t,a}$ are respectively the normal and abnormal subsets, we hypothesize that there exists a feature extractor $\psi:~\mathbb{R}^{h\times w \times c} \rightarrow \mathcal{X}$ that generates from $\mathcal{D}^t$ a compact dominant cluster $\mathcal{C}~\in~\mathcal{X}$ predominated by normal samples. 
\end{hypothesis*}
\noindent \emph{The anomaly scarcity assumption} often holds as it reflects most real-world scenarios where anomalies are rare compared to normal instances. In summary, our main objective is, therefore, to find or learn a feature extractor that verifies \emph{the dominant cluster existence hypothesis}.  This hypothesis is a core component of the proposed method discussed in Section~\ref{sec:proposed_method}, as it allows the introduction of a novel paradigm to approach UDA for one-class UAD consisting of (1) finding a feature exactor $\psi$ that can generate a compact dominant cluster of features $\mathcal{C}$ corresponding to normal samples within an unlabeled target dataset $\mathcal{D}^{\text{t}}$, (2) identifying the subset of samples $\mathcal{\Tilde{D}}^{\text{t,n}}$ corresponding to this cluster in the feature space of $\psi$, and (3) aligning the identified subset $\mathcal{\Tilde{D}}^{\text{t,n}}$ with the source normal samples $\mathcal{D}^{s,n}$ in the feature space of the source feature extractor $f$. Formally, we revisit Eq~(\ref{eq:previous_works_uda_bound}) as follows, 
\begin{equation}
    \epsilon^{t,n} \leq \epsilon^{s,n} + d(f(\mathcal{D}^{s,n}), f(\mathcal{\Tilde{D}}^{t,n})) + \lambda \ , 
    \label{eq:uda_bound}
\end{equation}
\noindent where $\mathcal{\Tilde{D}}^{t,n}= \{\mathbf X_i^t~|~\psi(\mathbf X_i^t)~\in~\mathcal{C} \}$.  Note that $\psi$ can be obtained by focusing on learning compact cross-domain features from which $\mathcal{C}$ can be identified through feature grouping and selection techniques such as clustering or filtering. As such, the proposed paradigm for UDA in one-class UAD lays the foundation for future research, where various technical choices can be explored at each stage.
\section{Proposed Solution to the Two-fold Unsupervised Curse}

\label{sec:proposed_method}
\begin{figure*}[t]
    \centering
    \includegraphics[scale=0.3]{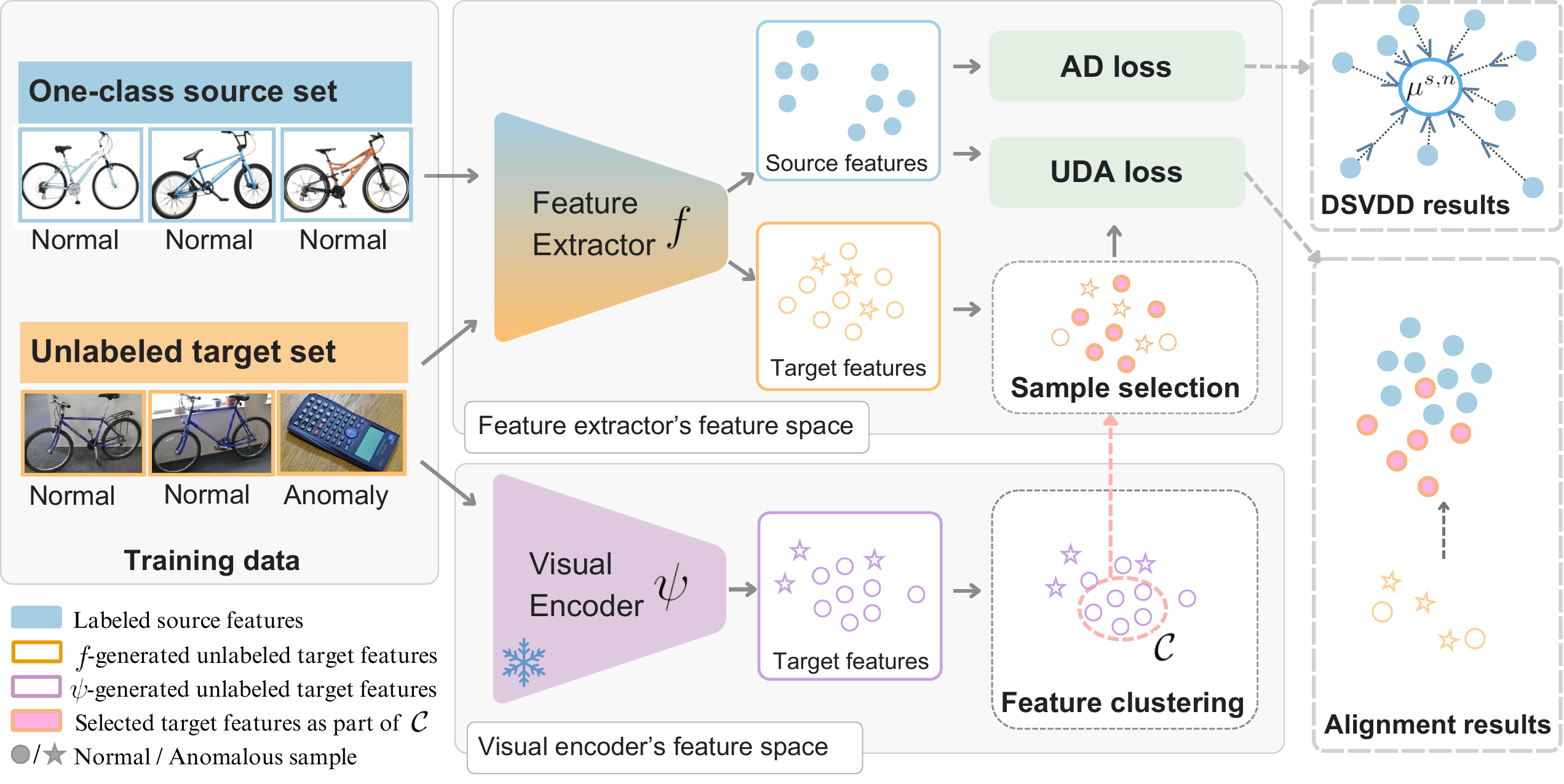}
    \caption{\textbf{Overview of the proposed method}: The top branch uses a trainable feature extractor with a DSVDD objective for one-class source data. The bottom branch clusters the features using a frozen CLIP visual encoder to identify the dominant feature cluster and align it with normal source representations. \raisebox{-1.25ex}{\textcolor{black}{\huge\textbullet}} and \textcolor{black}{$\bigstar$} denote normal and anomalous samples respectively.}
    \label{fig:main_figure}
\end{figure*}
Building on the assumption and hypothesis formulated in~\Cref{sec:generic_framework}, we present the technical choices, implemented as one possible solution for addressing UDA for semantic one-class-based image UAD, as illustrated in~\Cref{fig:main_figure}. 

Our approach has two branches. The upper branch depicts a trainable backbone $f$ that learns from both source and target domain data. The source features are optimized using a Deep Support Vector Data Description (DSVDD) objective~\cite{dsvdd-2018}. The lower branch focuses on visual feature extraction from the unlabeled target domain, through a frozen CLIP visual encoder~\cite{clip-2021}, defined as the $\psi$ feature extractor. Clustering is applied to these visual features to estimate the dominant cluster $\mathcal{C}$. Samples identified within $\mathcal{C}$ in the $\psi$ visual encoder's representation space are then selected within the space of the feature extractor $f$ and then aligned with the normal source features. \textcolor{black}{An algorithm is provided in Section~\textcolor{cvprblue}{A} of the Supplementary.}\\
\textbf{Training.} Specifically, given source and target image datasets $\mathcal{D}^{s,n}$ and $\mathcal{D}^{t}$, we apply DSVDD on the source data, enforcing feature compactness by minimizing the radius of a hypersphere to encapsulate the normal source representations. This is done by solving the following optimization problem,
\vspace{-4mm}
\begin{equation} 
    \min_{\theta_{f}} \mathcal{L}_{AD} = \min_{\theta_{f}} \frac{1}{N_s} \sum_{i=1}^{N_s} \|f(\mathbf{X}_i^s) - \boldsymbol{\mu}^{s,n}\|_{2}^{2},  \forall \mathbf{X}_i^s \in \mathcal{D}^{s,n},
\end{equation}
where $\boldsymbol{\mu}^{s,n}$ is the mean of the source features. For clustering, we use a $K$-means algorithm. Note that $\psi$ can be $f$ itself or any frozen visual encoder such as CLIP~\cite{clip-2021} or DINO-v2~\cite{dinov2-2023}.
The dominant cluster is identified as,
\begin{equation}
    \mathcal{C} = \arg\max_{\mathcal{C}_k} \left| \mathcal{C}_k \right| \text{ for } \, k \in \{1,..., K\},
\end{equation}
where $|\mathcal{C}_k|$ is the size of the $k$-th cluster $\mathcal{C}_k$, and $K$ is a hyper-parameter defining the number of expected components in the space of $\psi(\mathcal{D}^t)$. When clustering is applied to $f(\mathcal{D}^t)$, the selected features for alignment are $\mathcal{\Tilde{D}}^{t,n} = \mathcal{C}$. When clustering is applied to $\psi(\mathcal{D}^t)$, the selected samples are:
\vspace{-1mm}
\begin{equation}
    \mathcal{\Tilde{D}}^{t,n} = \{f(\mathbf{X}_i^t) \mid \psi(\mathbf{X}_i^t) \in \mathcal{C}\}\text{ }\forall \mathbf{X}_i^t \in \mathcal{D}^t
\end{equation}
Alignment between source and target features is achieved using a contrastive strategy, where the loss of a single positive source-target pair $\ell_{i,j}$:
\vspace{-1mm}
\begin{equation}
    \ell_{i,j} = -\log \frac{\exp\left(\frac{1}{\tau} \cdot \text{sim}\left(f(\mathbf{X}_i^s), f(\mathbf{X}_j^t)\right)\right)}
    {\sum_{p=1}^{N_t} \mathbbm{1}_{[\mathbf{X}_p^t \notin \mathcal{\Tilde{D}}^{t,n}]} \exp\left(\frac{1}{\tau} \cdot\text{sim}\left(f(\mathbf{X}_i^s), f(\mathbf{X}_p^t)\right) \right)}
\end{equation}
where $\text{sim}(\cdot, \cdot)$ denotes the cosine similarity, and $\tau$ is a temperature hyper-parameter. The UDA loss is computed as:
\vspace{-2mm}
\begin{equation}
    \mathcal{L}_{UDA} = \frac{1}{N_{s} \times |\mathcal{\Tilde{D}}^{t,n}|} \sum_{i=1}^{N_{s}} \sum_{j=1}^{|\mathcal{\Tilde{D}}^{t,n}|} \ell_{i,j},
\end{equation}
Finally, the overall loss is given by:
\vspace{-1mm}
\begin{equation}
    \mathcal{L} = \lambda_1\text{ }\mathcal{L}_{AD} + \lambda_2\text{ }\mathcal{L}_{UDA},  
\end{equation}
where $\lambda_1$ and $\lambda_2$ are hyper-parameters for $\mathcal{L}_{AD}$ and $\mathcal{L}_{UDA}$.

\noindent \textbf{Inference.} Note that the visual encoder $\psi$ is discarded at inference and that only the feature extractor $f$ is used to determine whether the input data is anomalous by calculating whether it falls inside or outside the hypersphere estimated by the DSVDD model. \textcolor{black}{Our method's algorithm is given in Section~\textcolor{cvprblue}{A} of the Supplementary.}

\section{Experimental Results}
\label{sec:experiments}
\begin{table*}
\setlength{\tabcolsep}{3pt}
    \begin{subtable}[c]{.5\textwidth}
    \centering
    \resizebox{\textwidth}{!}{%
        \begin{tabular}{p{0.16\linewidth}|c|cccccc|c}
            \toprule
            \multicolumn{1}{c}{\multirow{2}{*}{\makecell{{Normal} \\ {class}}}} & \multicolumn{1}{c}{\multirow{2}{*}{\makecell{{Source only} \\ {DSVDD}}}} & \multicolumn{6}{c}{{Few-shot}} & \makecell{{Unsupervised}} \\ \cmidrule{3-8}
            \multicolumn{1}{c}{} & \multicolumn{1}{c}{} & BiOST & TSA & ILDR & {IRAD} & {MsRA} &  \multicolumn{1}{c}{\textbf{Ours}} &{\textbf{Ours}} \\ \midrule
            \multicolumn{9}{c}{ \textbf{Clip Art} $\rightarrow$ \textbf{Product (C $\rightarrow$ P)}} \\ \midrule
            
            {Bike}       & 97.48 & 43.00 & 69.10 & 89.90 & 90.30 & 94.30 & 98.34 & 85.71 \\
            {Calculator} & 83.47 & 69.00 & 72.20 & 84.90 & 82.20 & 98.70 & 97.76 & 97.70 \\
            {Drill}      & 81.57 & 66.40 & 66.20 & 75.30 & 73.00 & 84.50 & 74.19 & 96.64 \\
            {Hammer}     & 83.32 & 50.10 & 77.40 & 74.70 & 84.50 & 80.10 & 89.55 & 82.63 \\
            {Kettle}     & 87.74 & 63.00 & 63.10 & 77.50 & 75.80 & 85.50 & 94.08 & 89.16 \\
            {Knives}     & 78.09 & 48.80 & 51.90 & 55.20 & 63.90 & 64.40 & 79.25 & 76.63 \\
            {Pan}        & 74.00 & 57.70 & 63.70 & 72.20 & 76.00 & 80.50 & 93.08 & 91.07 \\
            {Paperclip}  & 53.04 & 27.40 & 74.70 & 78.70 & 67.40 & 79.70 & 71.18 & 67.98 \\ 
            {Scissors}   & 86.45 & 56.40 & 64.70 & 79.50 & 68.90 & 85.50 & 87.71 & 88.43 \\
            {Soda}       & 51.21 & 50.20 & 57.40 & 70.30 & 53.30 & 72.40 & 61.16 & 92.37 \\ \midrule
            
            \small{Avg.} & \normalsize{77.64} & 53.20 & 66.04 & 75.82 & \normalsize{73.53} & {82.56} & \underline{84.63} & \normalsize{\textbf{86.83}} \\
            \small{$\pm$std} & \small{$\pm$14.04} & \small{$\pm$11.65} & \small{$\pm$7.36} & \small{$\pm$8.81} & \small{$\pm$10.24} & \small{$\pm$9.33} & \small{$\pm$11.93} & \small{$\pm$8.66}\\
            \midrule
            \multicolumn{9}{c}{ \textbf{Product} $\rightarrow$ \textbf{Clip Art (P $\rightarrow$ C)}}\\ \midrule
            {Bike}       & 82.55 & 52.70 & 65.80 & 83.10 & 85.70 & 86.60 & 82.06 & 92.99 \\
            {Calculator} & 62.82 & 65.20 & 63.40 & 87.20 & 79.20 & 91.90 & 91.59 & 89.88 \\ 
            {Drill}      & 71.81 & 47.00 & 57.10 & 63.90 & 71.20 & 73.50 & 70.58 & 77.54 \\
            {Hammer}     & 68.02 & 43.70 & 68.60 & 60.20 & 77.00 & 73.00 & 84.33 & 65.42 \\
            {Kettle}     & 71.85 & 47.70 & 61.50 & 68.80 & 70.00 & 73.40 & 75.38 & 78.19 \\ 
            {Knives}     & 57.22 & 63.10 & 57.50 & 65.30 & 70.30 & 73.10 & 77.74 & 71.99 \\ 
            {Pan}        & 71.44 & 49.30 & 63.50 & 69.30 & 72.80 & 80.00 & 83.72 & 82.46 \\ 
            {Paperclip}  & 26.19 & 45.10 & 49.90 & 69.70 & 61.80 & 69.00 & 67.05 & 55.93 \\
            {Scissors}   & 63.42 & 38.60 & 70.10 & 66.20 & 70.00 & 72.30 & 86.35 & 77.63 \\ 
            {Soda}       & 66.82 & 56.90 & 55.80 & 60.20 & 63.29 & 59.40 & 69.08 & 62.63 \\ \midrule
            
            \small{Avg.} & 64.21 & 50.93 & 61.32 & 69.39 & 72.13 & 75.22 & \textbf{78.79} & \underline{75.47} \\
            
            \small{$\pm$std} & \small{$\pm$14.22} & \small{$\pm$8.11} & \small{$\pm$5.94} & \small{$\pm$8.55} & \small{$\pm$6.76} & \small{$\pm$8.62} & \small{$\pm$7.74} & \small{$\pm$11.13}\\
            \bottomrule
            
            \end{tabular}
    }
    \caption{\textbf{OfficeHome~\cite{officehome}}}
    \end{subtable}
    \begin{subtable}[c]{.5\textwidth}
        \centering
        \resizebox{\textwidth}{!}{%
        \begin{tabular}{p{0.2\linewidth}|c|cccccc|c}
            \toprule
            \multicolumn{1}{c}{\multirow{2}{*}{\makecell{{Normal} \\ {class}}}} & \multicolumn{1}{c}{\multirow{2}{*}{\makecell{{Source only} \\ {DSVDD}}}} & \multicolumn{6}{c}{{Few-shot}} & \makecell{{Unsupervised}} \\ \cmidrule{3-8}
            \multicolumn{1}{c}{} & \multicolumn{1}{c}{} & BiOST & TSA & ILDR & {IRAD} & {MsRA} &  \multicolumn{1}{c}{\textbf{Ours}} &{\textbf{Ours}} \\ \midrule
            \multicolumn{9}{c}{ \textbf{Webcam} $\rightarrow$ \textbf{Amazon (W $\rightarrow$ A)}} \\ \midrule
            Backpack        & 86.48 & 59.90 & 76.30 & 91.90 & 90.20 & 95.20 & 95.40 & 97.62 \\ 
            Bookcase        & 35.77 & 56.60 & 59.60 & 78.40 & 82.20 & 84.50 & 76.25 & 91.16 \\
            Bottle          & 70.00 & 60.80 & 66.80 & 74.50 & 72.10 & 74.00 & 72.48 & 77.32 \\
            Desk Chair      & 56.92 & 57.60 & 63.40 & 85.30 & 80.90 & 87.20 & 85.50 & 92.06 \\ 
            Desk Lamp       & 82.26 & 50.50 & 60.90 & 72.60 & 67.50 & 70.00 & 82.38 & 81.50 \\
            Headphones      & 88.91 & 57.60 & 75.90 & 88.90 & 81.60 & 92.20 & 92.53 & 95.06 \\
            Keyboard        & 79.83 & 58.20 & 69.90 & 88.30 & 93.20 & 95.40 & 95.40 & 93.36 \\
            Laptop          & 51.79 & 59.10 & 63.00 & 86.20 & 98.10 & 99.00 & 95.63 & 79.97 \\
            Mouse           & 83.95 & 65.80 & 53.40 & 84.90 & 79.60 & 89.90 & 96.65 & 92.97 \\ 
            Pen             & 48.54 & 68.50 & 69.10 & 75.50 & 71.40 & 73.90 & 72.72 & 71.20 \\ \midrule
            
            \small{Avg.}    & 68.45 & 59.46 & 65.83 & 82.65 & 81.68 & 86.13 & \underline{86.49} & \textbf{87.22}\\
            
            \small{$\pm$std} & \small{$\pm$17.84} & \small{$\pm$4.70}& \small{$\pm$6.86}& \small{$\pm$6.46}& \small{$\pm$9.37} & \small{$\pm$9.72} & \small{$\pm$9.44} & \small{$\pm$8.48}\\
            \midrule
            
            \multicolumn{9}{c}{ \textbf{Amazon} $\rightarrow$ \textbf{Webcam (A $\rightarrow$ W)}} \\ \midrule
            Backpack        & 79.42 & 47.90 & 59.00 & 81.60 & 91.20 & 97.50 & 99.28 & 97.59 \\
            Bookcase        & 60.68 & 49.90 & 72.30 & 88.90 & 89.40 & 93.10 & 85.23 & 94.29 \\ 
            Bottle          & 40.94 & 66.00 & 69.80 & 86.90 & 95.30 & 96.20 & 93.65 & 94.95 \\
            Desk Chair      & 71.66 & 67.00 & 66.20 & 76.10 & 90.30 & 90.10 & 93.67 & 99.08 \\
            Desk Lamp       & 94.63 & 55.50 & 68.60 & 73.10 & 81.30 & 83.90 & 94.57 & 97.61 \\
            Headphones      & 70.99 & 68.30 & 72.40 & 93.70 & 91.60 & 96.00 & 96.54 & 96.04 \\
            Keyboard        & 77.90 & 66.00 & 76.90 & 91.10 & 95.70 & 98.10 & 90.62 & 76.59 \\
            Laptop          & 91.61 & 62.10 & 72.20 & 85.70 & 97.10 & 98.20 & 94.32 & 97.67 \\ 
            Mouse           & 72.17 & 69.10 & 69.40 & 82.20 & 85.40 & 86.50 & 96.35 & 81.41 \\ 
            Pen             & 44.26 & 79.10 & 86.10 & 97.60 & 98.90 & 99.60 & 97.09 & 99.99 \\ \midrule
            
            \small{Avg.}    & 70.43 & 63.09 & 71.29 & 85.69 & 91.62 & \underline{93.92} & \textbf{94.13} & 93.52\\ 
            
            \small{$\pm$std}& \small{$\pm$16.81} & \small{$\pm$9.03}& \small{$\pm$6.66}& \small{$\pm$7.26} & \small{$\pm$5.15} & \small{$\pm$5.11} & \small{$\pm$3.72} & \small{$\pm$7.52}\\
            \bottomrule
            
        \end{tabular}%
    }
    \caption{\textbf{Office31~\cite{office31}}}
    \end{subtable}
    \vspace{-3mm}
    \caption{Ten-run average and standard deviation of AUC (\%) on the Office datasets against previous SoA.}
    \vspace{-2mm}
    \label{tab:main_results_office}
\end{table*}
\begin{table}[h!]
\centering
\scalebox{0.7}{%
\begin{tabular}{p{0.95cm}|c|ccc|c}
\toprule
\multicolumn{1}{c}{\multirow{2}{*}{\makecell{{Normal}\\{class}}}} & \multicolumn{1}{c}{\multirow{2}{*}{\makecell{\small{Source only} \\ \small{(DSVDD)}}}} & \multicolumn{3}{c}{{Few-shot}} & \makecell{{Unsup.}} \\ \cmidrule{3-5}
 \multicolumn{1}{c}{} & \multicolumn{1}{c}{} & BiOST & {MsRA} & \multicolumn{1}{c}{\textbf{Ours}} & \textbf{Ours}\\ \midrule
\multicolumn{6}{c}{ \large \textbf{Synthetic} (CAD) $\rightarrow$ \textbf{Real}} \\ \midrule

Aero.        & 67.71 & 36.80 & 81.56 & 81.55 &  84.86 \\
Bicycle      & 65.12 & 59.20 & 68.45 & 74.58 &  81.45 \\
Bus          & 66.01 & 47.90 & 68.12 & 72.26 &  82.17 \\
Car          & 78.65 & 53.80 & 69.44 & 82.78 &  62.76 \\
Horse        & 67.24 & 58.00 & 68.77 & 80.17 &  83.52 \\ 
Knife        & 62.43 & 54.10 & 70.39 & 71.52 &  68.82 \\ 
Motor.       & 69.45 & 58.10 & 65.64 & 80.16 &  91.15 \\ 
Person       & 42.11 & 58.70 & 59.18 & 51.24 &  69.68 \\ 
Plant        & 57.77 & 42.10 & 65.81 & 71.46 &  70.58 \\ 
Skate.       & 60.70 & 41.60 & 61.30 & 63.17 &  83.71 \\ 
Train        & 54.75 & 52.40 & 69.73 & 60.62 &  69.98 \\ 
Truck        & 62.08 & 43.10 & 59.05 & 73.67 &  57.84 \\ \midrule

\small{Avg.} &  62.84 & 50.48 & 67.28 & \underline{71.93} & \normalsize{\textbf{75.54}}\\
\small{$\pm$std} &  \small{$\pm$8.55} & \small{$\pm$7.55}& \small{$\pm$5.79} & \small{$\pm$9.08} & \small{$\pm$ 9.80}\\
\bottomrule

\end{tabular}%
}
\caption{Ten-run average and standard deviation of AUC (\%) on the on the VisDA dataset~\cite{visda-2017} against previous SoA.}
\vspace{-3mm}
\label{tab:main_results_visda}
\end{table}

\subsection{Experimental Setting}
\label{sec:experimental_setting}
This section describes the datasets and the baselines used and the implementation details of our experiments. We report the performance using Area Under the ROC Curve (AUC) using \textbf{bold} and \underline{underline} for the best and second performances, respectively. The supplementary materials provide further details on each subsection. 
\vspace{-4mm}
\paragraph{Datasets.}
We evaluate our approach on four standard UDA benchmark datasets, \textbf{Office-Home}~\cite{office31}, \textbf{Office31}~\cite{office31},\textbf{VisDA}~\cite{visda-2017}, and \textbf{PACS}~\cite{pacs-2017}. \textcolor{black}{The types of domain shift of each dataset is described in Section~\textcolor{cvprblue}{A} of the Supplementary.} For the AD task, we adopt a standard one-vs-all protocol, where a single class is considered normal and the remaining are anomalies. We adopt the experimental protocol of previous DA works~\cite{msra-2021,irad-2023,biost-2019} to allow for a fair comparison --that is, we show results on ten classes from the ClipArt and Product domains for Office-Home, ten classes from Webcam and Amazon for Office31, and twelve classes from \textcolor{black}{the domains of Computer Aided Designs (CAD)  (synthetic objects) and real object photos} of VisDA. On PACS, like~\cite{ad-shift-2023}, we consider the Photo domain as source and the remaining three domains as targets.
\vspace{-0.35cm}
\paragraph{Baselines.} 
As no other works on UDA for semantic image UAD were previously introduced, we compare our method with several few-shot SoA approaches. More specifically, we consider \textbf{BiOST}~\cite{biost-2019} which is a one-shot approach and \textbf{TSA}~\cite{tsa-2021}, \textbf{ILDR}~\cite{ildr-2019}, \textbf{IRAD}~\cite{irad-2023}, and \textbf{MsRA}~\cite{msra-2021} that are few-shot methods. Furthermore, we introduce a few-shot variant (\textbf{Ours-Few-shot}) of our approach, which augments the target domain with normal and pseudo-anomalous samples similar to~\cite{jigsaw-neg-pairs-2021}. This augmentation yields semantically positive and negative pairs~\cite{jigsaw-neg-pairs-2021}, which are useful for the contrastive alignment strategy described in~\Cref{sec:proposed_method}. \textcolor{black}{Further details are given in Section~\textcolor{cvprblue}{A} of the Supplementary.}
\vspace{-0.75cm}
\paragraph{\textcolor{black}{Implementation details.}}In all experiments, the source set has only normal data, while the unlabeled target set includes mostly normals with 10\% randomly sampled anomalies. Training uses SGD with a cosine-annealing scheduler, learning rate of $10^{-3}$, weight decay of $5 \times 10^{-7}$, batch size 256 and $\lambda_1$ and $\lambda_2$ are set to 1. CLIP-ViT-B32 is the frozen visual encoder $\psi$ for feature clustering. To align with the setting of the baselines~\cite{msra-2021,irad-2023,biost-2019,tsa-2021}, ResNet50 is the trainable backbone $f$, initialized on ImageNet~\cite{imagenet-2009}. K-means clustering~\cite{kmeans-1979} uses 2, 10, and 5 components for Office, VisDA, and PACS, respectively. Like the baselines, the few-shot experiments use 10 (Office, PACS) and 100 shots (VisDA) \textbf{labeled as normal}, respectively.

\subsection{Comparison against State-of-the-art.}
\label{sec:sota}
Our method outperforms previous SoA on all benchmarks of our evaluation, as shown in~\Cref{tab:main_results_office} and~\Cref{tab:main_results_visda}. More specifically, our fully unsupervised UDA-UAD importantly improves upon previous few-shot SoA on C~$\rightarrow$~P and W~$\rightarrow$~A of the Office-Home~\cite{officehome} and Office31~\cite{office31} datasets. In addition, we observe an improvement of over $10\%$ in the VisDA dataset~\cite{visda-2017} with the fully unsupervised methodology over previous few-shot approaches despite being challenged by the two-fold unsupervised curse. These results highlight the relevance of the proposed method, even in the presence of a large domain gap, as in the case of synthetic CAD images and real-world photos.

In the P~$\rightarrow$~C and A~$\rightarrow$~W adaptation of the Office datasets, our few-shot variant also registers SoA performance, closely followed by our model trained under the fully unsupervised setting. These results highlight the flexibility of our framework, which can leverage minimal labeled target data when available but remains highly effective in a fully unsupervised setting.

Furthermore, we compare the performance of our model to two {pretrained visual encoders}~\cite{clip-2021}, namely \textcolor{black}{ResNet50} and CLIP-ViT-B32 in~\Cref{tab:adaptation_paradigms}. While the CLIP-ViT-B32 architecture achieves an average AUC of 72.08\%, our unsupervised method (75.54\%) still outperforms it on the VisDA dataset~\cite{visda-2017}. In contrast, the ResNet50 shows a significantly lower performance, with an average AUC of only 53.45\%. These results demonstrate that despite their strong performance, pretrained visual encoders are not specifically tailored for the domain adaptation task; thus, they remain vulnerable to domain shift. Therefore, training domain adaptation-specific models is still necessary to effectively bridge the gap between two given domains.

\subsection{Additional Experiments}
\label{sec:ablations}
\begin{wrapfigure}{r}{0.5\linewidth} 
    \centering
    \includegraphics[width=\linewidth]{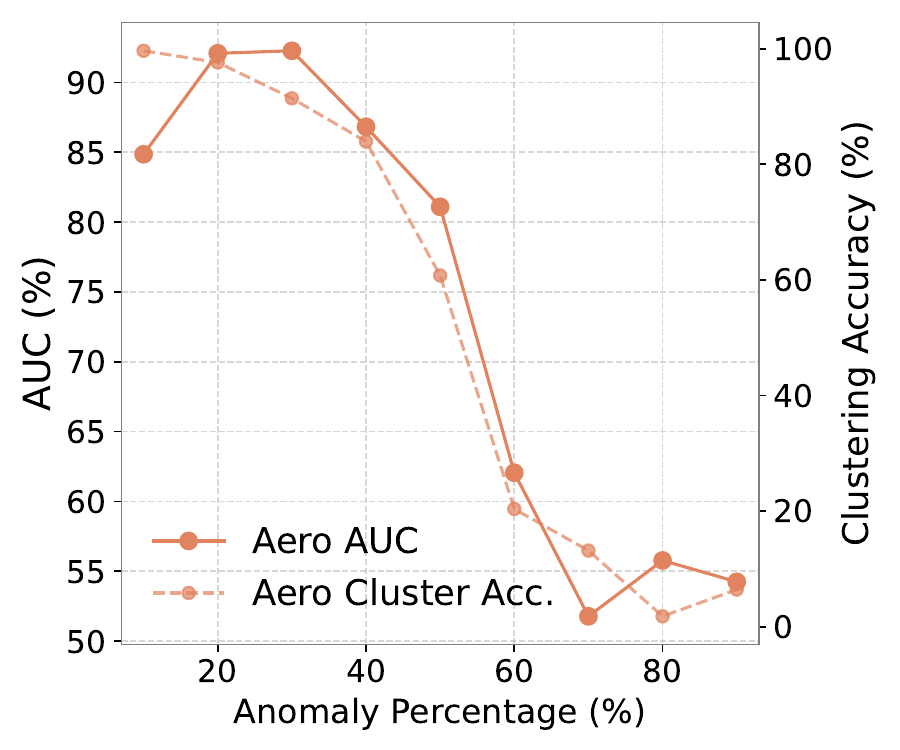}
    \vspace{-7mm}
    \caption{Assessing the validity of anomaly scarcity assumption.}
    \vspace{-4mm}
    \label{fig:cluster_acc_anomaly_percentage}
\end{wrapfigure}

Unless stated otherwise, all the following experiments are performed on VisDA~\cite{visda-2017}. Additional results on different components of our method are given in the Supplementary.
\vspace{-8mm}
\paragraph{\textcolor{black}{Anomaly scarcity assumption.}}To evaluate the impact of anomaly scarcity, we vary the anomaly ratio in the unlabeled target set from 10\% to 90\% and report our method's performance alongside clustering accuracy in~\Cref{fig:cluster_acc_anomaly_percentage} for the Aeroplane class from VisDA. The results indicate a strong correlation between AUC performance and clustering accuracy. As the anomaly proportion increases, the AUC gradually degrades, with a drastic drop beyond 50\%, where the dominant cluster assumption no longer holds. This is further evidenced by a significant decrease in the clustering accuracy. 
\vspace{-7mm}
\paragraph{Few-shot versus unsupervised paradigms.} The results presented in~\Cref{tab:adaptation_paradigms} compare pretrained visual encoders and source-only detectors with different adaptation paradigms, i.e., few-shot, unsupervised, and supervised (oracle). The source-only finetuned model improves slightly over the pretrained \textcolor{black}{ResNet50} visual encoder~\cite{resnet-2016} but still has lower performance than the adaptation approaches, achieving an average AUC of 62.84\%. Among the few-shot methods, our few-shot variant outperforms BiOST~\cite{biost-2019} and MsRA~\cite{msra-2021}, achieving the highest AUC of 71.93\%, which is comparable to the performance of a pretrained CLIP-ViT-B32 visual encoder. However, our unsupervised adaptation method surpasses all these models, with an average AUC of 75.54\% indicating its ability to effectively mitigate domain gaps without relying on labeled target data. This can be explained by the fact that after clustering, our model has access to more representative normal target data than few-shot models, hence better generalizing to the target normal class.
On the other hand, the Oracle, which has access to the target labels, achieves the highest performance (81.65\%). The small gap between our unsupervised method and the oracle demonstrates the effectiveness of our approach even without supervision.
\vspace{-4mm}
\begin{table}[h!]
\centering
\scalebox{0.69}{
\begin{tabular}{p{0.95cm}|p{0.8cm}p{0.8cm}|c|p{0.8cm}p{0.8cm}p{0.8cm}|p{0.8cm}|p{0.8cm}}
\toprule
\multicolumn{1}{c}{\multirow{3}{*}{\makecell{Normal\\class}}} & \multicolumn{2}{c}{Zero-shot} & \multicolumn{1}{c}{\multirow{3}{*}{\makecell{\small{DSVDD}\\\small{src fine-}\\\small{tuned only}}}} & \multicolumn{3}{c}{Few-shot Adaptation} & \multicolumn{1}{c}{\multirow{3}{*}{\makecell{\small Unsup.\\ \small Adapt.\\ \small \textbf{Ours}}}} & \multirow{3}{*}{\makecell{{\small Super-}\\{\small{(Adap.}}\\vised)}} \\ 
\cmidrule(lr){2-3} \cmidrule(lr){5-7}
 & R50 & CLIP &  & BiOST & MsRA & Ours &  &  \\ 
 \addlinespace[5pt]
\midrule
\multicolumn{9}{c}{\textbf{CAD} $\rightarrow$ \textbf{Real}} \\ \midrule

Aero.      & 41.05 & 74.97 & 67.71 & 36.80 & 81.56 & 81.55 & 84.86 & 90.91 \\
Bicycle    & 67.35 & 90.28 & 65.12 & 59.20 & 68.45 & 74.58 & 81.45 & 81.73 \\
Bus        & 28.58 & 42.27 & 66.01 & 47.90 & 68.12 & 72.26 & 82.17 & 72.16 \\
Car        & 32.48 & 64.16 & 78.65 & 53.80 & 69.44 & 82.78 & 62.76 & 68.42 \\
Horse      & 68.81 & 75.48 & 67.24 & 58.00 & 68.77 & 80.17 & 83.52 & 88.70 \\
Knife      & 67.78 & 95.28 & 62.43 & 54.10 & 70.39 & 71.52 & 68.82 & 78.90 \\ 
Motor.     & 60.07 & 82.25 & 69.45 & 58.10 & 65.64 & 80.16 & 91.15 & 83.46 \\ 
Person     & 71.69 & 56.26 & 42.11 & 58.70 & 59.18 & 51.24 & 69.68 & 85.19 \\
Plant      & 62.47 & 89.65 & 57.77 & 42.10 & 65.81 & 71.46 & 70.58 & 82.63 \\
Skate.     & 85.00 & 91.52 & 60.70 & 41.60 & 61.30 & 63.17 & 83.71 & 83.73 \\
Train      & 30.13 & 57.74 & 54.75 & 52.40 & 69.73 & 60.62 & 69.98 & 85.11 \\ 
Truck      & 26.05 & 45.08 & 62.08 & 43.10 & 59.05 & 73.67 & 57.84 & 78.91 \\ \midrule

{\small Avg.} & 53.45 & 72.08 & 62.84 & 50.48 & 67.28 & 71.93 & \underline{75.54} & \textbf{81.65} \\ 
{\small $\pm$std} & {\small $\pm$19.57} & {\small $\pm$17.83} & {\small $\pm$8.55} & {\small $\pm$7.55} & {\small $\pm$5.79} & {\small $\pm$9.08} & {\small $\pm$9.80} & {\small $\pm$6.11} \\
\bottomrule
\end{tabular}
}
\caption{AUC (\%) performance on the target domain of our UDA anomaly detector on VisDA~\cite{visda-2017} compared with various adaptation paradigms (from zero-shot, i.e., pretrained Visual encoders, few-shot, to supervised, i.e., Oracle).}
\label{tab:adaptation_paradigms}
\end{table}

\begin{table}[h!]
\centering
\scalebox{0.8}{
\begin{tabular}{c|c|c|c}
\toprule
 \small{\makecell{w/ Adaptation}} & \small{\makecell{w/ Clustering}} &  \small{\makecell{w/ CLIP $\psi$}} & \small{\textbf{AUC} (\%)} \\
\midrule
 
\xmark & \xmark & \xmark & 62.84$\pm$8.55\\

\cmark & \xmark & \xmark & 64.33$\pm$6.42\\

\cmark & \cmark & \xmark & 68.47$\pm$8.30\\

\cmark & \cmark & \cmark & \textbf{75.54}$\pm$9.80\\
\hline
\end{tabular}
}
\vspace{-1mm}
\caption{Ablation on the components of the proposed method.}
\vspace{-6mm}
\label{tab:ablation_all_components}
\end{table}

\paragraph{Ablation on the framework components.}
\label{sec:adaptation_paradigms} 
~\Cref{tab:ablation_all_components} provides the results obtained when each component, namely the adaptation loss, the dominant cluster identification through clustering, the use of an auxiliary visual encoder $\psi(\mathcal{D}^t)$ or the trainable features $f(\mathcal{D}^t)$. The results show that without adaptation, a model trained only on source data generalizes poorly to the target domain with only 62.84\%. Direct adaptation of the source and the unlabeled target without clustering leads to inconsistent results, indicating low generalization capabilities to the target domain. Introducing clustering results in a significant performance boost. This can be seen when clustering is applied to the original representations of the feature extractor, as the performance improves by +5.63\%, highlighting the importance of identifying the dominant cluster prior to alignment. Note that our method still outperforms the best few-shot baseline MsRA~\cite{msra-2021} \textcolor{black}{(68.47\% vs 67.28\%) with just clustering and alignment (\ie w/o CLIP)} across all the VisDA classes. Finally, the best results are achieved when all components are combined. This setup boosts the average AUC to 75.54\% on all VisDA classes. The substantial performance gains can be attributed to CLIP's rich visual features, which, together with clustering and alignment, help achieve a more robust anomaly detector capable of better handling domain shift.
\vspace{-1mm}

\begin{figure}[h]
    \centering
    \begin{minipage}{0.44\linewidth}
        \centering
        \includegraphics[width=\linewidth]{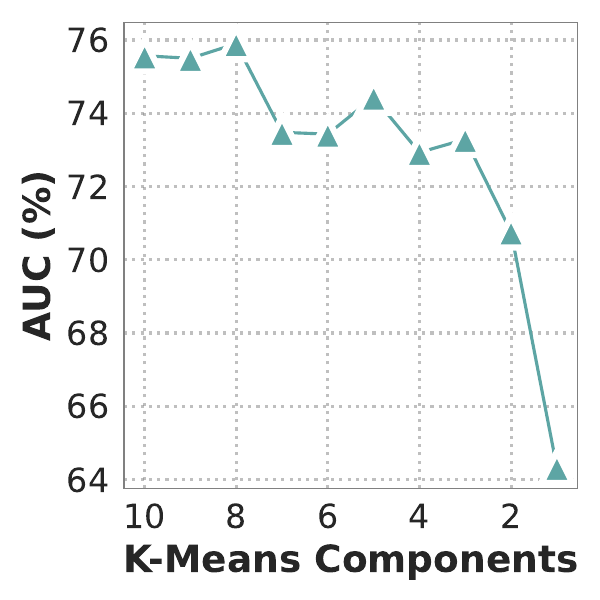}
        \vspace{-5mm}
        \caption{K-Means\cite{kmeans-1979}~components variation on VisDA.}
        \label{fig:auc_vs_kmeans}
    \end{minipage}
    \hfill
    \begin{minipage}{0.53\linewidth}
        \centering
        \scalebox{0.64}{
        \begin{tabular}{p{0.5cm}p{0.6cm}p{0.6cm}p{0.5cm}p{0.6cm}p{0.5cm}}
        \toprule
        \multicolumn{6}{c}{\small {$f$: ResNet50 + $\psi$: CLIP-ViT-B32}}\\ \midrule
        \multicolumn{3}{c}{\textbf{DSVDD~\cite{dsvdd-2018}}} & \multicolumn{3}{c}{\textbf{MSC~\cite{cmeanshift-2023}}} \\
        \cmidrule(lr){1-3} \cmidrule(lr){4-6}
        \textbf{ZS} & \textbf{Src} & \textbf{UDA} 
        & \textbf{ZS} & \textbf{Src} & \textbf{UDA} \\
        \midrule
        53.45 & 62.84 & 75.54 & 74.39 & 72.87 & 81.10 \\
              &      & \multicolumn{1}{l}{(\textbf{+12.7$\uparrow$})} &  & & \multicolumn{1}{l}{(\textbf{+8.23$\uparrow$})} \\
        \bottomrule
        \end{tabular}
        }
        \vspace{-2mm}
        \captionof{table}{Our UDA approach on two anomaly detection methods: DSVDD~\cite{dsvdd-2018} and Mean-shifted Contrastive Learning (MSC)~\cite{cmeanshift-2023}. ZS and Src mean Zero-shot and Source only, respectively.}
        \label{tab:dsvdd_vs_cmeanshift}
    \end{minipage}
\end{figure}

\begin{table}[h!]
    \centering
    \small{
    \scalebox{0.70}{
    \begin{tabular}{c|c|cccc} \toprule
    \multicolumn{1}{c}{\multirow{2}{*}{Dataset}} & \multirow{2}{*}{\makecell{w/o Adapt.\\(Src Only)}} & \multicolumn{4}{c}{{w/ Adaptation}}\\ \cmidrule(lr){3-6}
    \multicolumn{1}{c}{}& & \multicolumn{1}{c}{Kmeans} & \multicolumn{1}{c}{GMM} & Meanshift & kNN \\ \midrule
    VisDA & 62.84$\pm$08.55 & \textbf{75.54}$\pm$10.23 &  72.24$\pm$08.81 & 74.14$\pm$07.44 & 71.65$\pm$06.68\\  \midrule
    A$\rightarrow$W & 72.57$\pm$18.69 &	 94.82$\pm$07.52 &  \textbf{96.45}$\pm$05.43 &  87.32$\pm$12.53 & 82.73$\pm$10.96 \\  
    W$\rightarrow$A & 67.70$\pm$18.32 &  \textbf{87.72}$\pm$12.63 &  87.00$\pm$13.60 &  86.68$\pm$08.53 & 83.63$\pm$06.04\\ \midrule
    C$\rightarrow$P & 77.31$\pm$15.13 &  90.54$\pm$14.06 &  \textbf{90.85}$\pm$11.18 &  85.95$\pm$10.94 & 78.67$\pm$15.19\\
    P$\rightarrow$C & 63.88$\pm$15.60 &  70.92$\pm$11.36 &  \textbf{76.15}$\pm$13.32 &  73.50$\pm$13.49 & 71.38$\pm$11.71\\ 
    \bottomrule
    \end{tabular}
    }}
    \vspace{-2mm}
    \caption{Ablation on different clustering techniques. GMM and K-means use $10$ components for VisDA and $2$ for other datasets. k=2 for VisDA and k=1 for the remaining datasets.}
    \vspace{-7mm}
    \label{tab:clusterin_algos}
\end{table}
\vspace{-2mm}
\paragraph{Clustering methods.} We compare different clustering techniques on three UDA benchmarks in~\Cref{tab:clusterin_algos}. The first observation we make is that any type of clustering improves the performance. K-means and GMM have comparable results, without one clearly and consistently outperforming the other across datasets and adaptation directions. Meanshift clustering offers a performance increase compared to source-only models. However, its performance remains lower than that of the other clustering methods. In our experiments, we chose K-means clustering as it achieves comparable performance to GMM while requiring fewer parameters and simpler optimization. We further investigate the optimal number of K-Means components, as shown in~\Cref{fig:auc_vs_kmeans}. The figure indicates that using 8 to 10 components yields the highest performance, with an AUC of approximately 75-76\%. Decreasing the number of components would gradually degrade the performance. This suggests that a lower number of clusters may not capture the characteristics of the majority class, leading to inaccurate clustering and thus negatively impacting the generalization of the anomaly detection model across domains.

\vspace{-8mm}

\paragraph{\textcolor{black}{Beyond DSVDD.}} To assess whether our alignment approach applies to other unsupervised AD methods, we replace DSVDD with Mean-shifted Contrastive loss (MSC)~\cite{cmeanshift-2023} in~\Cref{tab:dsvdd_vs_cmeanshift}. MSC adapts contrastive loss to the one-class setting by shifting augmented representations of the normal samples toward the mean of pretrained normal features, preserving their compactness. Similar to~\cite{panda-2021,red-pandas}, it uses a kNN density estimator to detect anomalies. Our results suggest that both methods benefit from the adaptation, as a consistent average improvement of +12.7\% and +8.23\% is seen across all the 12 classes of VisDA.
\vspace{-2mm}
\paragraph{Alignment strategies.}
By comparing several alignment strategies in~\Cref{tab:alignment_strategies}, we observe that any alignment strategy, in general, improves the performance consistently for all adaptation benchmarks. Contrastive alignment consistently outperforms other adaptation losses, including statistical (MMD)~\cite{mmd-2006} and adversarial (GRL)~\cite{grl-2015} strategies.

\begin{table}[h]
    \centering
    \small{
    \scalebox{0.83}{
    \begin{tabular}{c|c|c|c|c} \toprule
    Dataset & Source Only & GRL~\cite{grl-2015} & MMD~\cite{mmd-2006} & Contrastive~\cite{infonce-loss-2018} \\ \midrule
    VisDA & 62.84$\pm$08.55 & 71.84$\pm$10.89 &  73.12$\pm$10.36 & \textbf{75.54}$\pm$9.80\\  \midrule
    A$\rightarrow$W & 72.57$\pm$18.69 &	{90.43}$\pm$10.68 & 86.86$\pm$12.97 & \textbf{94.82}$\pm$07.52 \\  
    W$\rightarrow$A &  67.70$\pm$18.32 &  83.49$\pm$11.04 &  {83.92}$\pm$09.90 & \textbf{87.72}$\pm$12.63\\ \midrule
    C$\rightarrow$P & 77.31$\pm$15.13 & 83.40$\pm$12.35 & 82.10$\pm$12.16 & \textbf{90.54}$\pm$14.06 \\
    P$\rightarrow$C & 63.88$\pm$15.60 &  66.78$\pm$15.02&  67.00$\pm$14.41& \textbf{70.92}$\pm$11.35\\ \bottomrule 
    \end{tabular}
    }}
    \vspace{-2mm}
    \caption{Performance in terms of AUC ($\%$) using different adaptation losses on the three UDA benchmarks.}
    \label{tab:alignment_strategies}
\end{table}

\paragraph{Comparison against domain generalization methods.}~\Cref{tab:pacs_dg_methods} compares the results of GNL~\cite{ad-shift-2023} with our proposed UDA method on the PACS dataset~\cite{pacs-2017}, with Photo as the source and Art, Cartoon, and Sketch as the target domains. It can be seen that our UDA approach consistently outperforms GNL~\cite{ad-shift-2023}, particularly on Cartoon and Sketch domains. This suggests that \textcolor{black}{unlike  DG methods, which aim to generalize to any unseen domain solely by training on the source domain, UDA can be more effective} for semantic UAD since it exposes the model to the target domain during training, even if it is unlabeled.
\vspace{-1mm}

\begin{table}[h!]
\centering
\scalebox{0.70}{
\begin{tabular}{llccc}
\toprule
Adapt. Type & Method & \small{Ph. $\rightarrow$ Art}    & \small{Ph. $\rightarrow$ Cartoon}    & \small{Ph. $\rightarrow$ Sketch}  \\
\midrule
None & \multicolumn{1}{l}{Source only} & 64.06 & 64.08 & 57.35 \\ \midrule 
DG   & GNL~\cite{ad-shift-2023} & 65.62 & 67.96 & \underline{62.39} \\  \midrule 
\multirow{2}{*}{DA} & MsRA~\cite{msra-2021} \small{(Few-shot)} & \textbf{71.43} & \underline{69.89} & 61.87\\ 
     & Ours (Unsup.)   & \underline{67.20} & \textbf{75.35} & \textbf{74.04} \\ 
\bottomrule
\end{tabular}
}
\vspace{-3mm}
\caption{AUC (\%) of Domain Generalization (DG) for anomaly detection, trained \emph{\textbf{ONLY}} on the source domain Photo (Ph.) and tested on unseen domains. DA means Domain Adaptation.}
\vspace{-5mm}
\label{tab:pacs_dg_methods}
\end{table}

\section{Limitations and Future Work}
\label{sec:limitations}
Our method, presented in Section~\ref{sec:proposed_method}, is one possible solution for addressing the problem of UDA for UAD. However, it is worth noting that it was tested in the context of semantic anomaly detection~\cite{semantic-ad-2024}, adopting a one-vs-all protocol, to facilitate the comparison with the closest baselines, namely~\cite{msra-2021,irad-2023}. These methods typically require the use of global features in contrast to standard anomaly detection, where fine-grained representations are usually targeted. For that reason, our method focuses mostly on global representations, while local features would be conceptually more suitable for fine-grained anomaly detection. In future works, we aim to extend our study to fine-grained anomaly detection by exploiting more relevant local representations.

\section{Conclusion}
\label{sec:conclusion}
This work is the first to address unsupervised domain adaptation (UDA) for one-class-based unsupervised anomaly detection (UAD), subject to what we refer to as the two-fold unsupervised curse. To address this ill-posed problem, an inherent property of anomalies, namely, their scarcity, is leveraged. This characteristic allows utilizing clustering, --as one possible solution-- for identifying a dominant cluster within the unlabeled target set. Assuming this cluster to be predominantly composed of normal data, a contrastive alignment strategy is then used to align its features with the normal source representations. Extensive experiments on standard UDA benchmarks demonstrate that the proposed method effectively mitigates the domain gap and enhances anomaly detection performance across different domains, outperforming other supervised adaptation approaches without requiring target annotations. Finding the optimal feature extractor remains an open research question. In future work, we intend to further explore compact representations across domains to improve the proposed domain adaptation framework.

\section{Acknowledgments}
This work was supported by the Fonds National de la Recherche (FNR) under the grant agreements no. BRIDGES2021/IS/16353350/FaKeDeTeR and UNFAKE, ref. 16763798, and by Post Luxembourg.
{
    \small
    \bibliographystyle{ieeenat_fullname}
    \bibliography{main}
}
\end{document}